\def\year{2018}\relax
\documentclass[letterpaper]{article} 
\usepackage{aaai18}  
\usepackage{times}  
\usepackage{helvet}  
\usepackage{courier}  
\usepackage{url}  
\usepackage{graphicx}  

\usepackage{booktabs}       
\usepackage{amsfonts}       
\usepackage{nicefrac}       
\usepackage{microtype}      
\usepackage{amsmath}
\usepackage{color}
\usepackage{adjustbox} 

\usepackage{array,multirow,graphicx}
\usepackage{rotating}

\begin{document}
%
 
\title{Acquiring Common Sense Spatial Knowledge through Implicit Spatial Templates}

\author{Guillem Collell \\
	Department of Computer Science\\
	KU Leuven \\
	gcollell@kuleuven.be
	\And
	Luc Van Gool \\
	Computer Vision Laboratory\\
	ETH Zurich \\
	vangool@vision.ee.ethz.ch
	\And
	Marie-Francine Moens \\
	Department of Computer Science\\
	KU Leuven \\
	sien.moens@cs.kuleuven.be	
	}

\maketitle

\begin{abstract}
Spatial understanding is a fundamental problem with wide-reaching real-world applications. The representation of spatial knowledge is often modeled with \textit{spatial templates}, i.e., regions of acceptability of two objects under an \textit{explicit} spatial relationship (e.g., ``on", ``below", etc.). In contrast with prior work that restricts spatial templates to \textit{explicit} spatial prepositions (e.g., ``glass \textit{on} table"), here we extend this concept to \textit{implicit} spatial language, i.e., those relationships (generally actions) for which the spatial arrangement of the objects is only implicitly implied (e.g., ``man \textit{riding} horse"). In contrast with \textit{explicit} relationships, predicting spatial arrangements from \textit{implicit} spatial language requires significant common sense spatial understanding. Here, we introduce the task of predicting spatial templates for two objects under a relationship, which can be seen as a spatial question-answering task with a (2D) continuous output (``where is the man w.r.t. a horse when the man is walking the horse?"). We present two simple neural-based models that leverage annotated images and structured text to learn this task. The good performance of these models reveals that spatial locations are to a large extent predictable from \textit{implicit} spatial language. Crucially, the models attain similar performance in a challenging \textit{generalized} setting, where the object-relation-object combinations (e.g.,``man walking dog") have never been seen before. Next, we go one step further by presenting the models with unseen objects (e.g., ``dog"). In this scenario, we show that leveraging word embeddings enables the models to output accurate spatial predictions, proving that the models acquire solid common sense spatial knowledge allowing for such generalization. 
\end{abstract}

\section{Introduction}

To provide machines with common sense is one of the major long term goals of artificial intelligence. Common sense knowledge regards knowledge that humans have acquired through a lifetime of experiences. It is crucial in language understanding because a lot of content needed for correct understanding is not expressed explicitly but resides in the mind of communicator and audience. In addition, humans rely on their common sense knowledge when performing a variety of tasks including interpreting images, navigation and reasoning, to name a few. Representing and understanding spatial knowledge are in fact imperative for any agent (human, animal or robot) that navigates in a physical world. In this paper, we are interested in acquiring spatial commonsense knowledge from language paired with visual data.

Computational and cognitive models often handle spatial representations as \textit{spatial templates} or regions of acceptability for two objects under an \textit{explicit} (a.k.a. \textit{deictic}) spatial preposition such as ``on", ``below" or ``left" \cite{logan1996computational}. Contrary to previous work that conceives spatial templates only for \textit{explicit} spatial language \cite{malinowski2014pooling,moratz2006spatial,logan1996computational}, we extend such concept to \textit{implicit} (a.k.a. \textit{intrinsic}) spatial language, i.e., relationships---generally actions---that do \emph{not} explicitly define the relative spatial configuration between the two objects (e.g., ``glass \textit{on} table") but only implicitly (e.g., ``woman \textit{riding} horse"). In other words, \textit{implicit} spatial templates capture the common sense spatial knowledge that humans possess and is not explicit in the language utterances. 

Predicting spatial templates for \textit{implicit} relationships is notably more challenging than for \textit{explicit} relationships. Firstly, whereas there are only a few tens of \textit{explicit} spatial prepositions, there exist thousands of actions, entailing thus a drastic increase in the sparsity of (\textit{object}$_1$, \textit{relationship}, \textit{object}$_2$) combinations. Secondly, the complexity of the task radically increases in \textit{implicit} language. More precisely, while \textit{explicit} spatial prepositions\footnote{Some prepositions (e.g., ``on") might be ambiguous as they can express other circumstantial arguments such as time. Here, we refer to spatial prepositions once they have been disambiguated as such.}
are highly deterministic about the spatial arrangements (e.g., (\textit{object}$_1$, below, \textit{object}$_2$) unequivocally implies that \textit{object}$_1$ is relatively lower than \textit{object}$_2$), actions generally are not. E.g., the relative spatial configuration of ``man" and the object is clearly distinct in (man, pulling, kite) than in (man, pulling, luggage) yet the action is the same. Contrarily, other relationships such as ``jumping" are highly informative about the spatial template, i.e., in (\textit{object}$_1$, jumping, \textit{object}$_2$), \textit{object}$_2$ is in a lower position than \textit{object}$_1$. Hence, unlike \textit{explicit} relationships, predicting spatial layouts from \textit{implicit} spatial language requires spatial common sense knowledge about the objects, actions and their interaction, which suggests the need of learning to compose the triplet (Subject, Relationship, Object) as a whole instead of learning a template for each Relationship.

To systematically study these questions, we propose the task of predicting the relative spatial locations of two objects given a structured text input (Subject, Relationship, Object). We introduce two simple neural-based models trained from annotated images that successfully address the two challenges of \textit{implicit} spatial language discussed above. Our quantitative evaluation reveals that spatial templates can be reliably predicted from \textit{implicit} spatial language---as accurately as from \textit{explicit} spatial language. We also show that our models generalize well to templates of unseen combinations, e.g., predicting (man, riding, elephant) without having been exposed to such scene before, tackling thus the challenge of sparsity. Furthermore, by leveraging word embeddings, the models can correctly generalize to spatial templates with unseen words, e.g., predicting (man, riding, elephant) without having ever seen an ``elephant" before. Since word embeddings capture attributes of objects \cite{collell2016is}, one can reasonably expect that embeddings are informative about the spatial behavior of objects, i.e., their likelihood of exhibiting certain spatial patterns with respect to other objects. For instance, without having ever seen ``boots" before but only ``sandals", the model correctly predicts the template of (person, wearing, boots) by inferring that, since ``boots" are similar to ``sandals", they must be worn at the same location of the ``person"'s body. Hence, the model leverages the acquired common sense spatial knowledge to \textit{generalize} to unseen objects. Furthermore, we provide both, a qualitative 2D visualization of the predictions, and an analysis of the learned weights of the network which provide insight into the spatial connotations of words, revealing fundamental differences between \textit{implicit} and \textit{explicit} spatial language.

The rest of the paper is organized as follows. In Sect.~\ref{related_work} we review related research. In Sect.~\ref{deep_spatial_templates} we first introduce the task of predicting spatial templates and then present two simple neural models. Then, in Sect.~\ref{experimental_setup}, we describe our experimental setup. In Sect.~\ref{results} we present and discuss our results. Finally, in Sect.~\ref{conclusions} we summarize the contributions of this article.

\section{Related work}
\label{related_work}

Spatial processing has drawn significant attention from the cognitive \cite{logan1996computational} and artificial intelligence communities \cite{kruijff2007situated}. More specifically, spatial understanding is essential in tasks involving text-to-scene conversion such as robots' understanding of natural language commands \cite{guadarrama2013grounding,moratz2006spatial} or robot navigation.
\\
\\
\noindent{ \bf Spatial templates.}
Earlier approaches have predominantly considered rule-based spatial representations \cite{kruijff2007situated,moratz2006spatial}. 
In contrast, \citeauthor{malinowski2014pooling} \shortcite{malinowski2014pooling} propose a learning-based pooling approach to retrieve images given queries of the form (object$_1$, spatial\_preposition, object$_2$). They learn the parameters of a \textit{spatial template} for each \textit{explicit} spatial preposition (e.g., ``left" or ``above") which computes a soft spatial fit of two objects under the relationship. E.g., an object to the left of the referent object obtains a high score for the ``left" template and low for the ``right" template. Contrary to them, we consider \textit{implicit} spatial language instead of \textit{explicit}. Additionally, while they build a spatial template for each (explicit) Relationship, we build a template for each (Subject, Relationship, Object) combination, allowing the template to be determined by the interaction/composition of the Subject, Relationship and Object instead of the Relationship alone. Additionally, the model from \citeauthor{malinowski2014pooling} \shortcite{malinowski2014pooling} does not output spatial arrangements of objects, nor can it perform predictions with \textit{generalized} (unseen) relationships.  
\\
\\
\noindent{ \bf Leveraging spatial knowledge in tasks.} It has been shown that knowledge of the spatial structure in images improves the task of image captioning \cite{elliott2013image}. These authors manually annotate images with geometric relationships between objects and show that a rule-based caption generation system benefits from this knowledge. In contrast to this work, our interest lies in predicting spatial arrangements of objects from text instead of generating text given images. 
Furthermore, while they employ a small domain of only 10 actions, our goal is to learn from frequent spatial configurations and generalize these to unseen and rare objects/actions (and their combinations). Spatial knowledge has also improved object recognition \cite{shiang2017vision}. These authors mine texts and labeled images to obtain spatial knowledge in the form of object co-occurrences and their relative positions. This knowledge is represented in a graph and a random walk algorithm over this graph results in a ranking of possible object labellings. Contrarily, our representations of spatial knowledge are neural network based, are not primarily used for object recognition, and we furthermore predict spatial templates.
\\
\\
\noindent{ \bf Common sense spatial knowledge.} \citeauthor{yatskar2016stating} \shortcite{yatskar2016stating} propose a model to extract common sense facts from annotated images and their textual descriptions using co-occurrence statistics among which is point-wise mutual information (PMI). These facts include six spatial relationships (``on", ``under", ``touches", "above", "besides", "holds", "on", and "disconnected"). The result is a symbolic (discretized) representation of common sense knowledge in the form of relations between objects that logically entail other relational facts. Our method also extracts common sense knowledge from images and text, but predicts (continuous) spatial templates. \citeauthor{lin2015don} \shortcite{lin2015don} leverage common sense visual knowledge (e.g., object locations and co-occurrences) in the tasks of fill-in-the-blank and visual paraphrasing. They compute the likelihood of a scene to identify the most likely answer to multiple-choice textual scene descriptions. In contrast, we focus solely on spatial information---and in assuring the correctness of our spatial predictions---rather than on scene understanding.
\\
\\  
\noindent{ \bf Image generation.} Although models that generate images from text exist (e.g., DRAW model \cite{gregor2015draw}), their focus is quite distant from producing ``spatially sensible" images and they are generally meant to generate a single object rather than placing it relative to other objects.
\\
\\
As discussed, spatial knowledge can improve a wide range of tasks \cite{shiang2017vision,lin2015don,elliott2013image}. This suggests that the predictions of our models can be used as spatial common sense input for methods that rely on good spatial priors. Additionally, existing methods require the spatial information to be present in the data and lack the capacity to extrapolate/generalize. Thus, in this paper we focus, first, on showing that \textit{generalizing} spatial arrangements to unseen objects and object-relation-object combinations is possible and, second, on ensuring that the predicted templates are accurate by performing several quantitative and qualitative evaluations.

\begin{figure}[t]
	\begin{center}
		\includegraphics[scale=0.45]{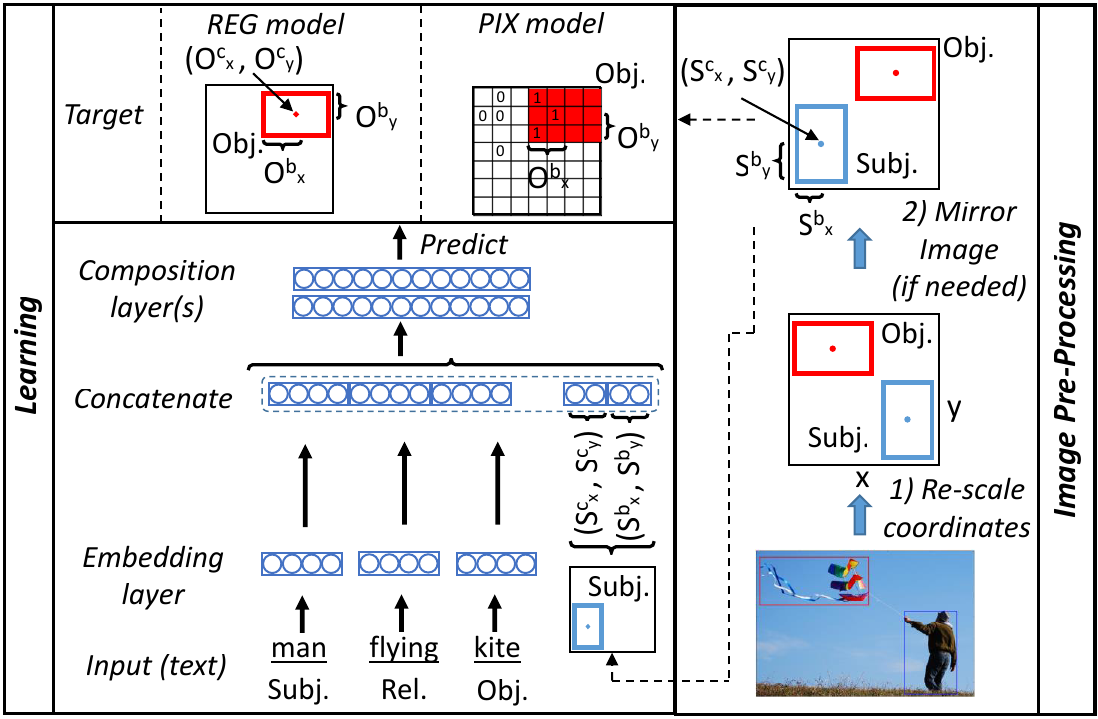}
	\end{center}
	\caption{Overview of our models (left) and the image pre-processing setting (right).}
	\label{model_scheme}
\end{figure}

\section{Proposed task and model}
\label{deep_spatial_templates}

\subsection{Proposed task}
\label{proposed_task}

To learn spatial templates, we propose the task of predicting the 2D relative spatial arrangement of two objects under a relationship given a structured text input of the form (Subject, Relationship, Object)---henceforth abbreviated as ($S$, $R$, $O$). Let us denote the 2D coordinates of the center (``$c$") of the Object's box as $O^{c} = [O^{c}_{x}, O^{c}_{y}] \in\mathbb{R}^2$, where $O^{c}_{x}\in\mathbb{R}$ and $O^{c}_{y}\in\mathbb{R}$ are the horizontal and vertical components respectively. Let $O^{b} = [O^{b}_{x}, O^{b}_{y}] \in\mathbb{R}^2$ be half of the width ($O^{b}_{x}$) and half of the height ($O^{b}_{y}$) of the Object's box (``$b$"). We employ a similar notation for the Subject ($S^{c},S^{b}$), and model predictions are denoted with a hat $\widehat{O^{c}}, \widehat{O^{b}}$. The task consists in predicting the Object's location and size $[O^{c}$,$O^{b}]\in\mathbb{R}^4$ (\textbf{output}) given the structured text \textbf{input} ($S$, $R$, $O$) and the location $S^{c}$ and size $S^{b}$ of the Subject (Fig.~\ref{model_scheme}).\footnote{Crucially, we notice that knowing the Subject's coordinates is not a requirement for generating templates (but only for evaluating against the ground truth) since inputting arbitrary coordinates (e.g., $S^{c}$=[0.5, 0.5]) still enables visualizing \emph{relative} object locations. Additionally, we input the Subject's size in order to provide a reference size to the model. However, we find that without this input, the model still learns to predict an ``average size" for each Object.} This defines a supervised task where the size and location of bounding boxes in images serve as ground truth. Our task can be interpreted as a spatial question-answering with structured questions (triplets) that allows evaluating the answer quantitatively in a 2D space. Hence, the goal is to answer \textit{common sense spatial questions} such as ``if a man is feeding a horse, \textit{where} is the man relative to the horse?" or "\textit{where} would a child wear her shoes?"

\subsection{Proposed models}
\label{proposed_model} 

To build a mapping from the \textit{input} to the \textit{output} in our task (Sect.~\ref{proposed_task}) we propose two simple neural models (Fig.~\ref{model_scheme}). Their architecture is identical in the input and representation layers, yet they differ in the output and loss function.

\paragraph{\textit{(i) Input and representation layers.}} An \textbf{embedding layer} maps the three \textbf{input} words ($S$, $R$, $O$) to their respective $d$-dimensional vectors $w_S W_{S}, w_R W_{R}, w_O W_{O}$, where $w_{S}\in\mathbb{R}^{|V_S|}, w_{R}\in\mathbb{R}^{|V_R|}, w_{O}\in\mathbb{R}^{|V_O|}$ are one-hot encodings of $S$, $R$, $O$ (a.k.a. one-of-$k$ encoding, i.e., a sparse vector with 0 everywhere except for a 1 at the position of the $k$-th word) and $W_{S}\in\mathbb{R}^{d\times|V_S|}, W_{R}\in\mathbb{R}^{d\times|V_R|}, W_{O}\in\mathbb{R}^{d\times|V_O|}$ their embedding matrices with $|V_S|, |V_R|, |V_O|$ their vocabulary sizes. This layer represents objects and relationships as continuous features, enabling thus to introduce external knowledge of unseen objects as features. The embeddings are then concatenated with the Subject center $S^{c}$ and size $S^{b}$ in a vector $[w_S W_{S}, w_R W_{R}, w_O W_{O}, S^{c}, S^{b}]$ which is inputted to a stack of hidden layers that \textbf{compose} $S$, $R$ and $O$ into a joint hidden representation $z_h$: 
\[ z_h = f(W_h[w_S W_{S}, w_R W_{R}, w_O W_{O}, S^{c}, S^{b}] + b_h) \] 
where $f(\cdot)$ is an element-wise non-linear function and $W_h$ and $b_h$ are the weight matrix and bias respectively.\footnote{Similarly, $z_h$ can be composed with more hidden layers.}

\paragraph{\textit{(ii) Output and loss.}} We consider two different possibilities for the output $\hat{y}$ that follows immediately after the last layer: $z_{out} = W_{out}z_h + b_{out}$ (Fig.~\ref{model_scheme}, top left).
\begin{itemize}
	\item \textbf{Model 1 (REG)}. A \textit{regression} model where the output are the Object coordinates and size $\hat{y}=z_{out}=[\widehat{O^{c}}$,$\widehat{O^{b}}]\in\mathbb{R}^4$ and is evaluated against the true $y=[O^{c}, O^{b}]$ with a mean squared error (MSE) loss: $L(y,\hat{y}) = \lVert \hat{y} - y\lVert^2 $.
	\item \textbf{Model 2 (PIX)}. The output $\hat{y}=\sigma(z_{out})=(\hat{y}_{i,j})\in\mathbb{R}^{M\times M}$, where $M$ is the number of pixels per side and $\sigma()$ an element-wise sigmoid, is now a 2D heatmap of \textit{pixel} activations $\hat{y}_{i,j}\in[0,1]$ indicating the probability that a pixel belongs to the Object (class 1). Predictions $\hat{y}$ are evaluated against the true $y=(y_{i,j})\in\mathbb{R}^{M\times M}$ with a binary cross-entropy loss: $L(y,\hat{y}) = -\sum_{i=1}^{M}\sum_{j=1}^{M}y_{i,j}\hat{y}_{i,j} - (1-y_{i,j})\log(1-\hat{y}_{i,j})$, where $y_{i,j}\in\{0,1\}$.
\end{itemize}

These models are conceptually different and have different capabilities. While the REG model outputs ``crisp" pointwise predictions, PIX can model more diffuse spatial templates where the location of the object has more variability, e.g., in (man, flying, kite) the ``kite" can easily move around.
Notice that in contrast with convolutional neural networks (CNNs) our approach does not make use of the image pixels (Fig.~\ref{model_scheme}), yielding a model fully specialized in spatial knowledge.

\section{Experimental setup}
\label{experimental_setup}

We employ a 10-fold cross-validation (CV) setting. Data are randomly split into 10 disjoint parts and 10\% is employed for testing and 90\% for training, repeating this for each of the 10 folds. Reported results are averages over the 10 folds. 

\subsection{Visual Genome data set}
\label{visual_genome}
\begin{figure}[!h]
	\begin{center}
		\includegraphics[scale=0.36]{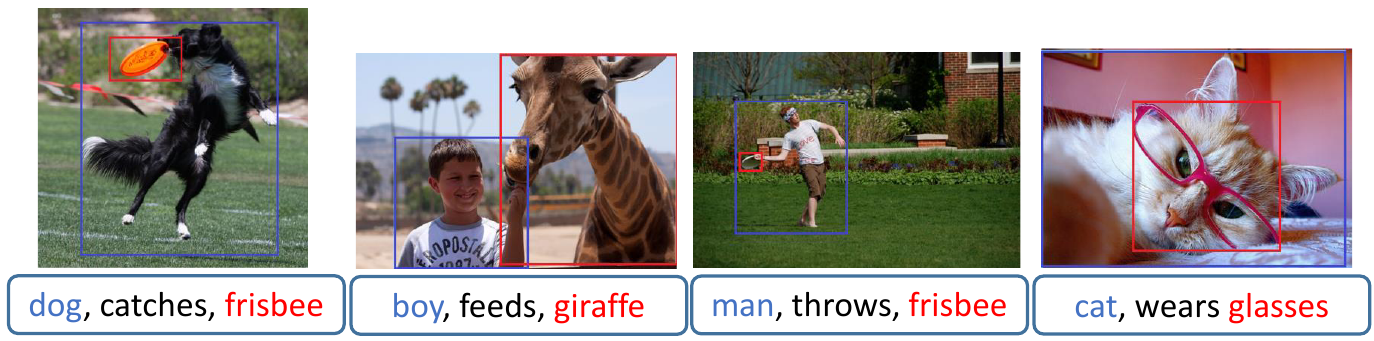}
	\end{center}
	\caption{Sample of images with object boxes and relationships from Visual Genome.}
	\label{vg_samples}
\end{figure}

We use the Visual Genome dataset \cite{krishnavisualgenome} as our source of annotated images. The Visual Genome consists of $\sim$108K images containing $\sim$1.5M human-annotated (Subject, Relationship, Object) instances with bounding boxes for Subject and Object (Fig.~\ref{vg_samples}). We filter out all the instances containing \textit{explicit} spatial prepositions, preserving only instances with \textit{implicit} spatial relationships. We keep only combinations for which we have word embeddings available for the whole triplet ($S$, $R$, $O$). After this filtering, $\sim$378K instances are preserved, yielding 2,183 unique \textit{implicit} Relationships and 5,614 unique objects (i.e., Subjects and Objects). The left out instances of \textit{explicit} language yield $\sim$852K instances, 36 unique \textit{explicit} spatial prepositions and 6,749 unique objects. 

\subsection{Evaluation sets}
\label{eval_sets}

We consider the following subsets of the Visual Genome data to evaluate performance.

\noindent{(i) \bf Raw} data: Simply the unfiltered instances from the Visual Genome data (Sect.~\ref{visual_genome}). This set contains a substantial proportion of meaningless (e.g., (nose, almost, touching)) and irrelevant (e.g., (sign, says, gate 2)) instances. 

\noindent{(ii) \bf Generalized Triplets}: We pick at random 100 combinations ($S$, $R$, $O$) among the 1,000 most frequent \textit{implicit} combinations in Visual Genome. This yields $\sim$25K instances such as (person, holding, racket), (man, flying, kite), etc.\footnote{This evaluation set along with our Supplementary material are available at https://github.com/gcollell/spatial-commonsense.}

\noindent{(iii) \bf Generalized Words}: We randomly choose 25 objects (e.g., ``woman", ``apple", etc.)\footnote{The complete list of objects is: [surfboard, shadow, head, surfer, woman, bear, bag, sunglasses, hair, apple, grass, water, eye, shoes, foot, jeans, jacket, bus, bike, cat, sky, elephant, tree, plane, eyes].} among the 100 most frequent objects in Visual Genome and take all the instances ($\sim$130K) that contain any of these words. For example, since ``apple" is in our list, e.g., (cat, sniffing, apple) is kept.\footnote{We also evaluated a list of \textit{generalized} Relationships, obtaining similar results. We additionally tested two extra lists of \textit{generalized} objects, which yielded consistent results.} Notice that a combination ($S$, $R$, $O$) with a \textit{generalized word} is automatically a \textit{generalized triplet} too. 

When enforcing \textbf{generalization} conditions in our experiments, all combinations from sets (ii) and (iii) are removed from the training data to prevent the model from seeing them.\footnote{To avoid confusion with the term `zero-shot' in classification, we denote our unseen words/triplets as \textit{generalized}. Although both settings have resemblances, they differ in that, in ours, the unseen categories are inputs while in classification are targets. Notice that in both settings one must necessarily have semantic knowledge about the ``zero-shot" class at hand \cite{socher2013zero} (e.g., in the form of word embeddings), otherwise the task is clearly infeasible.} Even without imposing \textit{generalization} conditions (or when testing with \textbf{Raw} data), reported results are always on unseen \textit{instances}---yet the \textit{combinations} ($S$, $R$, $O$) may have been seen during training (e.g., in different images). All sets above contain exclusively \textit{implicit} spatial language, although an analogous version of the \textit{Raw} set where $R$ are \textit{explicit} spatial prepositions is also considered in our experiments. 

\subsection{Data pre-processing}
\label{data_pre-processing}
The coordinates of bounding boxes in the images are normalized by the width and height of the image. Thus, $S^{c},O^{c}\in[0,1]^2$. Additionally, we notice that the distinction between left and right is arbitrary regarding the semantics of the image \cite{singhal2003probabilistic}. That is, a mirrored image preserves entirely its meaning---while a vertically inverted image does not. For example, a ``child" ``walking" a ``horse" can meaningfully be at either side of the ``horse", while a ``child" ``riding" a ``horse" cannot be either above or below the ``horse". Hence, to free the model from such arbitrariness, we \textbf{mirror} the image when (and only when) the Object is at the left hand side of the Subject. This leaves the Object always to the right-hand side of the Subject. Notice that mirroring is aimed at properly measuring performance and does not impose any big constraint (one can simply consider the symmetric reflection of the predictions as equally likely). 

\subsection{Evaluation metrics}
The REG model directly outputs Object coordinates, while PIX outputs 2D heatmaps. We however enable evaluating the PIX model with regression/classification metrics by taking the point of maximum activation (or their average, if there are many) as the Object center $\widehat{O^{c}}$. The predicted Object size $\widehat{O^{b}}$ is not estimated. We use the following performance metrics.

\paragraph{(A) Intersection over Union (IoU).} We compute the bounding box overlap (IoU) from the PASCAL VOC object detection task \cite{everingham2015pascal}: $\text{IoU}=\frac{area(\widehat{B_O}\cap B_O)}{area(\widehat{B_O}\cup B_O)}$ where $\widehat{B_O}$ and $B_O$ are the predicted and ground truth Object bounding boxes, respectively. \emph{If the IoU is larger than 50\%, the prediction is counted as correct.} It must be noted that our setting is not comparable to object detection (nor our results) since we do not employ the image as input (but text) and thus we cannot leverage the pixels to predict the Object's location. 

\paragraph{(B) Regression.} We consider standard regression metrics.

\noindent{(i) \bf Coefficient of Determination (R$^2$)} of model predictions $\hat{y}=[\widehat{O^{c}},\widehat{O^{b}}]$ and ground truth $y=[O^{c},O^{b}]$. R$^2$ is widely used to evaluate goodness of fit of a model in regression. The best R$^2$ score is 1 while the worst one is arbitrarily negative. A constant prediction would obtain a score of 0.

\noindent{(ii) \bf Pearson Correlation (r)} between the predicted $\widehat{O^{c}_{x}}$ and the true $O^{c}_{x}$ $x$-component of the Object center. Similarly for the $y$-components $\widehat{O^{c}_{y}}$ and $O^{c}_{y}$. 

\paragraph{(C) Above/below classification.} With the semantic distinction between vertical and horizontal axes in mind (Sect.~\ref{data_pre-processing}), we consider the problem of classifying above/below relative locations. If the model predicts that the Object center is above/below the Subject center and this actually occurs in the image we count it as correct. That is, $sign(\widehat{O^{c}_{y}} - S^{c}_{y})$ and $sign(O^{c}_{y} - S^{c}_{y})$ must match. We report \textbf{macro-averaged F1 (F1$_y$)} and \textbf{macro-averaged accuracy (acc$_y$)}.

\paragraph{(D) Pixel macro accuracy (px-acc).} Accuracy or correct classification rate per pixel may not be 
a good measure here, where class 1 (Object box) comprises, on average, only 5\% of the pixels. Thus, a constant prediction of zeros everywhere would obtain 95\% accuracy. Hence, we consider \textbf{macro-averaged pixel accuracy (px-acc)}, namely the average of recall of class 0 and recall of class 1.
We report the best px-acc across the full range of decision thresholds on the predicted pixel-wise confidence scores.

\subsection{Word embeddings}
We use 300-dimensional GloVe \textit{word embeddings} \cite{pennington2014glove} pre-trained on the Common Crawl corpus (consisting of 840B-tokens), which we obtain from the authors' website.\footnote{ http://nlp.stanford.edu/projects/glove}

\subsection{Model hyperparameters and implementation}
\label{model_parameters}
Our experiments are implemented in Python 2.7 and we use Keras deep learning framework for our models \cite{chollet2015keras}.
Model hyperparameters are first selected in a 10-fold cross-validation setting and we report (averaged) results on 10 new splits. Models are trained for 10 epochs on batches of size 64 with the RMSprop optimizer using a learning rate of 0.0001 and 2 hidden layers with 100 ReLu units. We find that the models are not very sensitive to parameter variations. The parameters of the embeddings are not backpropagated, although we find that this choice has little effect on the results. We employ a $15 \times 15$ pixel grid as output of the PIX model.

\begin{table}[t]
	\centering
	\scalebox{0.75}{
\begin{tabular}{@{}llccccccc@{}}
\toprule
 &  & R$^2$ & acc$_y$ & F1$_y$ & r$_x$ & r$_y$ & IoU & px-acc \\ \midrule
\multirow{8}{*}{\rotatebox[origin=c]{90}{Implicit}}  & REG$_\textit{EMB}$ & 0.704 & 0.751 & 0.750 & 0.892 & 0.835 & 0.117 & - \\
 & REG$_\textit{RND}$ & 0.693 & 0.750 & 0.749 & 0.890 & 0.828 & 0.120 & - \\
 & REG$_\textit{1H}$ & \textbf{0.720} & \textbf{0.764} & \textbf{0.764} & \textbf{0.897} & \textbf{0.843} & \textbf{0.152} & - \\
 & \textit{ctrl} & -0.999 & 0.522 & 0.522 & 0.002 & 0.001 & 0.075 & - \\ \cmidrule(l){2-9} 
 & PIX$_\textit{EMB}$ & - & 0.713 & 0.714 & 0.831 & 0.755 & - & \textbf{0.862} \\
 & PIX$_\textit{RND}$ & - & 0.702 & 0.702 & 0.821 & 0.742 & - & 0.856 \\
 & PIX$_\textit{1H}$ & - & \textbf{0.716} & \textbf{0.717} & \textbf{0.852} & \textbf{0.778} & - & 0.857 \\
 & \textit{ctrl} & - & 0.514 & 0.513 & 0.000 & -0.001 & - & 0.500 \\ \midrule
\multirow{8}{*}{\rotatebox[origin=c]{90}{Explicit}}  & REG$_\textit{EMB}$ & 0.585 & 0.771 & 0.773 & 0.810 & 0.822 & 0.345 & - \\
 & REG$_\textit{RND}$ & 0.576 & 0.769 & 0.769 & 0.806 & 0.815 & 0.327 & - \\
 & REG$_\textit{1H}$ & \textbf{0.604} & \textbf{0.779} & \textbf{0.780} & \textbf{0.814} & \textbf{0.827} & \textbf{0.384} & - \\
 & \textit{ctrl} & -1.001 & 0.633 & 0.630 & 0.000 & -0.001 & 0.042 & - \\ \cmidrule(l){2-9} 
 & PIX$_\textit{EMB}$ & - & \textbf{0.729} & \textbf{0.726} & \textbf{0.716} & \textbf{0.768} & - & \textbf{0.817} \\
 & PIX$_\textit{RND}$ & - & 0.721 & 0.719 & 0.709 & 0.758 & - & 0.812 \\
 & PIX$_\textit{1H}$ & - & 0.723 & 0.720 & 0.709 & 0.760 & - & 0.810 \\
 & \textit{ctrl} & - & 0.589 & 0.581 & 0.001 & 0.000 & - & 0.500 \\ \bottomrule
\end{tabular}
	} 
	\caption{Results on the \textbf{Raw} test data.}
	\label{tab_raw}
\end{table}

\begin{table*}[t]
	\centering
	\scalebox{0.75}{
\begin{tabular}{@{}llccccccccccccccc@{}}
\toprule
 &  & \multicolumn{7}{c}{Generalization} & \multicolumn{1}{l}{} & \multicolumn{7}{c}{No Generalization} \\ \cmidrule(lr){3-9} \cmidrule(l){11-17} 
 &  & R$^2$ & acc$_y$ & F1$_y$ & r$_x$ & r$_y$ & IoU & px-acc &  & R$^2$ & acc$_y$ & F1$_y$ & r$_x$ & r$_y$ & IoU & px-acc \\ \cmidrule(r){1-9} \cmidrule(l){11-17} 
  \multirow{8}{*}{\rotatebox[origin=c]{90}{Triplets}} & REG$_\textit{EMB}$ & 0.746 & 0.784 & 0.777 & 0.903 & 0.877 & 0.122 & - &  & 0.774 & 0.795 & 0.797 & 0.908 & 0.889 & 0.151 & - \\
 & REG$_\textit{RND}$ & 0.731 & 0.770 & 0.770 & 0.899 & 0.863 & 0.126 & - &  & 0.776 & 0.796 & 0.797 & 0.909 & 0.887 & 0.161 & - \\
 & REG$_\textit{1H}$ & \textbf{0.764} & \textbf{0.790} & \textbf{0.794} & \textbf{0.906} & \textbf{0.880} & \textbf{0.166} & - &  & \textbf{0.791} & \textbf{0.802} & \textbf{0.807} & \textbf{0.913} & \textbf{0.895} & \textbf{0.203} & - \\
 & \textit{ctrl} & -1.097 & 0.516 & 0.506 & 0.001 & 0.002 & 0.077 & - &  & -1.097 & 0.515 & 0.506 & -0.004 & -0.002 & 0.077 & - \\ \cmidrule(lr){2-9} \cmidrule(l){11-17} 
 & PIX$_\textit{EMB}$ & - & \textbf{0.751} & \textbf{0.760} & 0.835 & 0.813 & - & \textbf{0.878} &  & - & \textbf{0.756} & \textbf{0.766} & 0.852 & 0.824 & - & 0.886 \\
 & PIX$_\textit{RND}$ & - & 0.740 & 0.750 & 0.836 & 0.799 & - & 0.868 &  & - & 0.744 & 0.756 & 0.843 & 0.820 & - & 0.885 \\
 & PIX$_\textit{1H}$ & - & 0.738 & 0.752 & \textbf{0.868} & \textbf{0.830} & - & 0.874 &  & - & 0.748 & 0.761 & \textbf{0.876} & \textbf{0.842} & - & \textbf{0.887} \\
 & \textit{ctrl} & - & 0.509 & 0.502 & -0.002 & 0.001 & - & 0.500 &  & - & 0.511 & 0.503 & -0.009 & -0.001 & - & 0.501 \\ \midrule
  \multirow{8}{*}{\rotatebox[origin=c]{90}{Words}} & REG$_\textit{EMB}$ & \textbf{0.633$^*$} & \textbf{0.742$^*$} & \textbf{0.741$^*$} & \textbf{0.877$^*$} & \textbf{0.795$^*$} & \textbf{0.075$^*$} & - &  & 0.725 & 0.788 & 0.786 & 0.896 & 0.856 & 0.128 & - \\
 & REG$_\textit{RND}$ & 0.438 & 0.630 & 0.629 & 0.852 & 0.603 & 0.046 & - &  & 0.718 & 0.788 & 0.787 & 0.895 & 0.850 & 0.132 & - \\
 & REG$_\textit{1H}$ & 0.425 & 0.596 & 0.586 & 0.855 & 0.621 & 0.053 & - &  & \textbf{0.740} & \textbf{0.802} & \textbf{0.802} & \textbf{0.900} & \textbf{0.862} & \textbf{0.167} & - \\
 & \textit{ctrl} & -1.022 & 0.519 & 0.518 & 0.000 & 0.000 & \textbf{0.075} & - &  & -1.017 & 0.518 & 0.518 & 0.002 & -0.001 & 0.074 & - \\ \cmidrule(lr){2-9} \cmidrule(l){11-17} 
 & PIX$_\textit{EMB}$ & - & \textbf{0.717$^*$} & \textbf{0.719$^*$} & 0.802 & \textbf{0.722$^*$} & - & \textbf{0.835$^*$} &  & - & 0.757 & 0.760 & 0.832 & 0.784 & - & \textbf{0.870} \\
 & PIX$_\textit{RND}$ & - & 0.644 & 0.643 & 0.764 & 0.587 & - & 0.780 &  & - & 0.747 & 0.749 & 0.823 & 0.775 & - & 0.866 \\
 & PIX$_\textit{1H}$ & - & 0.591 & 0.590 & \textbf{0.813} & 0.573 & - & 0.764 &  & - & \textbf{0.760} & \textbf{0.763} & \textbf{0.852} & \textbf{0.799} & - & 0.866 \\
 & \textit{ctrl} & - & 0.512 & 0.511 & -0.001 & 0.000 & - & 0.500 &  & - & 0.511 & 0.511 & 0.002 & -0.001 & - & 0.500 \\ \bottomrule
\end{tabular}
	} 
	\caption{Results on generalized \textbf{triplets} (top) and generalized \textbf{words} (bottom) (see Sect.~\ref{eval_sets}). The tables on the right show results on the same sets without imposing generalization conditions, i.e., allowing to see all combinations/words during training.}
		\label{tab_ZS}
\end{table*}

\section{Results and discussion}
\label{results}

We consider the evaluation sets from Sect.~\ref{eval_sets} and the following variations of PIX and REG models (Sect.~\ref{proposed_model}). The subindex \textit{\textbf{EMB}} denotes a model that employs GloVe embeddings and \textit{\textbf{RND}} a model with embeddings randomly drawn from a dimension-wise normal distribution of mean ($\mu$) and standard deviation ($\sigma$) equal to those of the GloVe embeddings, preserving the original dimensionality ($d$=300). A third type employs one-hot vectors (\textit{\textbf{1H}}). We additionally consider a control method (\textit{\textbf{ctrl}}) that outputs random normal predictions of $\mu$ and $\sigma$ equal to the dimension-wise mean and standard deviation of the training targets. We test statistical significance with a Friedman rank test and post hoc Nemenyi tests on the results of the 10 folds. We indicate with an asterisk $^*$ in the tables when a method is significantly better than the rest (p $<$ 0.01) within the same model (PIX or REG).

\subsection{Evaluation with raw data}

Table~\ref{tab_raw} shows that all methods perform well considering the amount of noise present in the \textit{Raw} data. Especially noteworthy is the finding that relative locations can be predicted from \textit{implicit} spatial language approximately as accurately as from \textit{explicit} spatial language. Interestingly, unlike the other metrics, the IoU (which only counts a prediction as correct if the overlap between true and predicted boxes is larger than 50\%) is clearly higher in \textit{explicit} than in \textit{implicit} language, which suggests that \textit{implicit} templates exhibit more flexibility on the Object's location. Hence, ``blurrier" predictions such as those of PIX can be a good choice in some applications, e.g., computing the soft fit between images and templates to perform (spatially informed) image retrieval. We also observe that models with \textit{1H} embeddings tend to perform better, yet differences are generally only significant against \textit{RND}.

\subsection{Generalized evaluations}

Table~\ref{tab_ZS} shows that all models perform well on \textit{generalized} \textbf{triplets} (top left), remarkably closely to their performance without imposing generalization conditions (right). Again, \textit{1H} performs slightly better, yet only significantly better than \textit{RND} (p $<$ 0.005). Notably, the good performance of \textit{RND} on \textit{generalized} triplets evidences that the model does not rely on external knowledge (word embeddings) to predict unseen combinations. This ability of generalizing from frequent combinations ($S$, $R$, $O$) to rare/unseen ones is especially valuable given the sparsity of \textit{implicit} combinations and the impossibility of learning all of them from data.

Contrarily, larger performance differences are observed with \textit{generalized} \textbf{words} (Tab.~\ref{tab_ZS}, bottom left) where, as expected, \textit{EMB} outperforms \textit{RND} and \textit{1H} embeddings by a margin thanks to the transference of knowledge from word embeddings combined with the acquired spatial knowledge.  

\subsection{Qualitative evaluation (spatial templates)}
\begin{figure*}[!h]
	\begin{center}
		\includegraphics[scale=0.47]{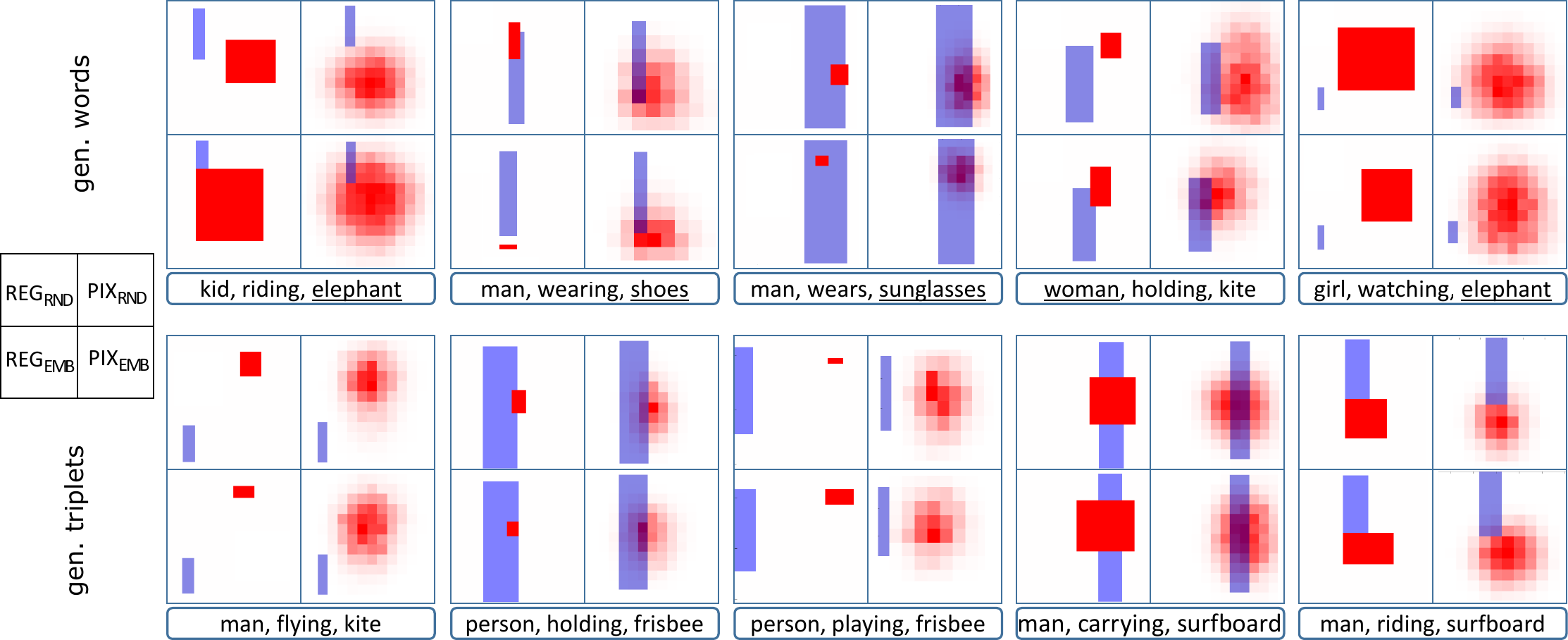}
	\end{center}
	\caption{Model predictions for \textit{generalized} words (\textbf{top}) and triplets (\textbf{bottom}). Model (PIX, REG) and embedding types (\textit{EMB}, \textit{RND}) are as indicated in the legend on the left. The Subject's location is given (\textcolor{blue}{\textbf{blue}} box) and the model predicts the Object (\textcolor{red}{\textbf{red}}). In PIX, the intensity of the red corresponds to the predicted probability. The \textit{generalized} (unseen) words are \underline{underlined}.}
	\label{quali_plot}
\end{figure*}

To ensure that model predictions are meaningful and interpretable, we further validate the quantitative results above with a qualitative evaluation. Notice that all plots in Fig.~\ref{quali_plot} are \textit{generalized} (either unseen \textit{words} or \textit{triplets}). Both, PIX and REG are able to infer the size and location of the Object notably well in unseen \textit{triplets} (Fig.~\ref{quali_plot}, bottom), regardless of the embedding type (\textit{EMB} or \textit{RND}). However, in unseen \textit{words} (Fig.~\ref{quali_plot}, top), the models that leverage word embeddings (\textit{EMB}) tend to perform better, aligning with the quantitative results above. Remarkably, both PIX$_\textit{EMB}$ and REG$_\textit{EMB}$ output very accurate predictions in generalized \textit{words} (top), e.g., predicting correctly the size (and location) of an ``elephant" relative to a ``kid" even though the model has never seen an ``elephant" before. 
Noticeably, the models learn to compose the triplets, distinguishing, for instance, between ``carrying a surfboard" and ``riding a surfboard" or between ``playing frisbee" and ``holding a frisbee", etc. 

\begin{figure}[t]
	\begin{center}
		\includegraphics[scale=0.5]{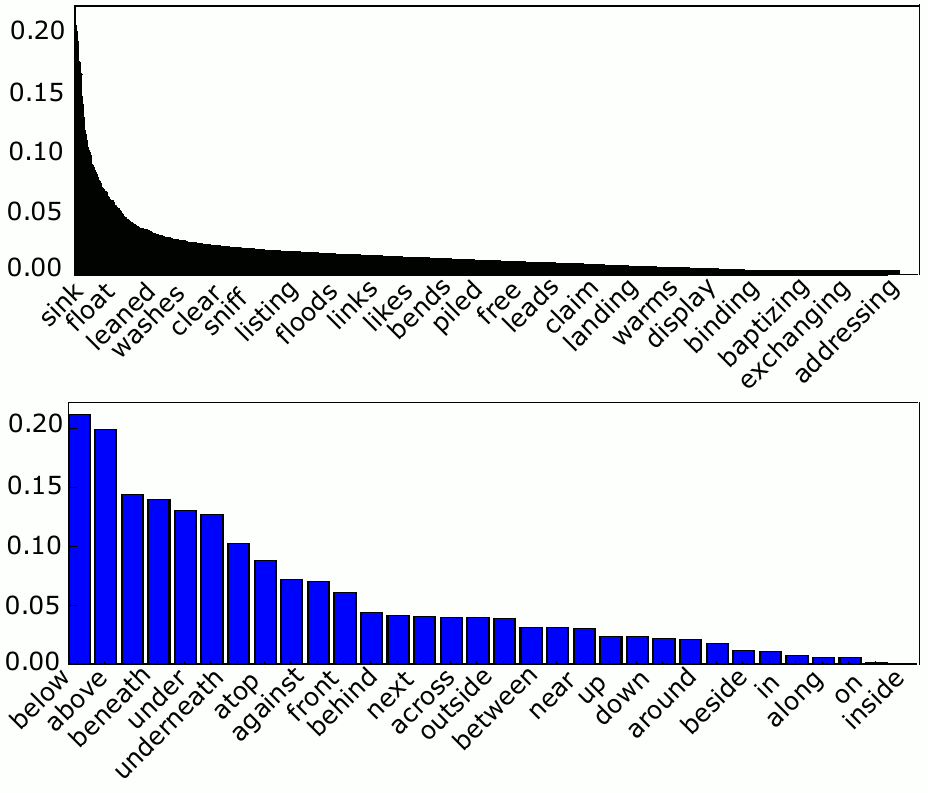}
	\end{center}
	\caption{Weights in absolute value of the REG model for \textit{implicit} (top) and \textit{explicit} (bottom) \textbf{Relationships}. For readability, the labels on the x-axis have been undersampled.}
	\label{weights_plot}
\end{figure}

The mapping of language to a 2D visualization provides another interesting property. Traditional language processing systems translate spatial information to qualitative symbolic representations that capture spatial knowledge with a limited symbolic vocabulary and that are used in qualitative spatial reasoning \cite{Cohn2008}. Research shows that translation of language to qualitative spatial symbolic representations is difficult \cite{Kordjamshidi2015}, obtaining rather low F1 measures on recognition performance, even if the language utterance is accompanied by visual data \cite{Kordjamshidi2017}. Here, we have shown that we can translate language utterances into visualizations in a quantitative 2D space, complementing thus existing symbolic models.

\subsection{Interpretation of model weights}
We study how the weights of the model provide insight into the spatial properties of words. To obtain more interpretable weights, we learn a REG\footnote{Unlike PIX, the REG model directly outputs coordinates and thus allows for an easy interpretation of the weights.} model without hidden layers, resulting in only an embedding layer followed by a linear output layer $\hat{y} = W_{out}u + b_{out}$, where $u:=[w_S, w_R, w_O, S^{c}, S^{b}]$. By using one-hot encodings $w_{S}, w_{R}, w_{O}$, the \textit{concatenation} layer $u$ becomes of size $|V_S| + |V_R| + |V_O| + 2 + 2$. E.g., if ``wearing" has one-hot index $j$ in the Relationships' vocabulary ($V_R$), its index in the concatenation layer is $|V_S| + j$. The product $W_{out}u$ is a 4-dimensional vector, where its $i$-th component is the product of the $i$-th row of $W_{out}$ with the vector $u$. Thus, the component $|V_S| + j$ of the $i$-th row of $W_{out}$ gives us the influence of the $j$-th relationship (i.e., ``wearing") on the $i$-th dimension of the output $\hat{y} = [\widehat{O_x^{c}},\widehat{O_y^{c}},\widehat{O_x^{b}},\widehat{O_y^{b}}]\in\mathbb{R}^4$.

\begin{table}[!h]
\centering
\resizebox{\columnwidth}{!}{ 
\Large
\begin{tabular}{@{}lclcllclc@{}}
\toprule
\multicolumn{4}{c}{Implicit} & \multicolumn{1}{c}{} & \multicolumn{4}{c}{Explicit} \\ \cmidrule(r){1-4} \cmidrule(l){6-9} 
\multicolumn{2}{c}{Objects} & \multicolumn{2}{c}{Relationships} & \multicolumn{1}{c}{} & \multicolumn{2}{c}{Objects} & \multicolumn{2}{c}{Relationships} \\ \cmidrule(r){1-4} \cmidrule(l){6-9} 
headband & -0.275 & flying & -0.194 &  & hook & -0.160 & below & -0.212 \\
visor & -0.266 & kicking & 0.148 &  & glasses & -0.160 & above & 0.199 \\
hoof & 0.246 & cutting & 0.142 &  & sailboat & -0.159 & beneath & -0.144 \\
sandals & 0.241 & catching & -0.132 &  & vase & 0.151 & under & -0.140 \\
kite & -0.234 & riding & 0.119 &  & woods & -0.149 & over & 0.131 \\ \cmidrule(r){1-4} \cmidrule(l){6-9}
fryer & 3.7e-5 & see & -6.8e-5 &  & spools & 7.5e-5 & in & -0.011 \\
books & 3.2e-5 & float & 5.8e-5 &  & glasss & -7.1e-5 & along & 0.008 \\
avenue & -3.0e-5 & finding & 5.2e-5 &  & dune & 7.0e-5 & at & -0.006 \\
english & 2.6e-5 & pulled & 2.6e-5 &  & cookies & 6.7e-5 & on & 0.005 \\
burger & 8.2e-5 & removes & 1.6e-5 &  & sill & -5.1e-5 & inside & -0.002 \\ \bottomrule
\end{tabular}
} 
\caption{Words with the ten largest (top) and smallest (bottom) weights in absolute value for the Object's $y$-coordinate ($\widehat{O_y^{c}}$), in \textit{implicit} and \textit{explicit} language (learned in \textit{Raw} data).}
\label{tab_weights}
\end{table}

Table~\ref{tab_weights} shows the weights influencing the $y$-coordinate $\widehat{O_y^{c}}$. We notice that objects such as ``kite" or ``headband" which tend to be above the Subject have a large negative weight, while objects that tend to be below, e.g., ``sandals" or ``hoof", have a large positive weight, i.e., a large positive influence on the Object's $y$-coordinate. While \textit{implicit} relations such as ``kicking" or ``riding" are strong predictors of the Object's $y$-coordinate, ``finding" or ``pulled" are weak, suggesting that the spatial template rather depends on their composition with the Subject and Object. In fact, the weights of \textit{implicit} relations such as ``flying" or ``riding" are comparable to those of \textit{explicit} relations such as ``above" or ``atop", behaving similarly to \textit{explicit} language. Notice that even the less informative \textit{explicit} relations have weights of at least one order of magnitude larger that the least informative \textit{implicit} relations. Figure~\ref{weights_plot} further evidences that \textit{explicit} relationships generally have larger weights than those of \textit{implicit} relationships. Altogether, the generally small weights of the implicit relations (and therefore their influence on the template) emphasize the need of composing the triplet (Subject, Relationship, Object) as a whole rather than modeling the Relationship alone. 

\section{Conclusions}
\label{conclusions}
	
Overall, this paper provides insight into the fundamental differences between \textit{implicit} and \textit{explicit} spatial language and extends the concept of \textit{spatial templates} to \textit{implicit} spatial language, the understanding of which requires common sense spatial knowledge about objects and actions. 
We define the task of predicting relative spatial arrangements of two objects under a relationship and present two embedding-based neural models that attain promising performance, proving that spatial templates can be accurately predicted from \textit{implicit} spatial language. Remarkably, our models generalize well, predicting correctly unseen (\textit{generalized}) object-relationship-object combinations. Furthermore, the acquired common sense spatial knowledge---aided with word embeddings---allows the model to correctly predict templates for unseen words. Finally, we show that the weights of the model provide great insight into the spatial connotations of words.

A first limitation of our approach is the fully supervised setting where the models are trained using images with detected ground truth objects and parsed text---which aims at keeping the design clean in this first study on \textit{implicit} spatial templates. Notice however that methods to automatically parse images and text exist. In future work, we aim at implementing our approach in a weakly supervised setting. A second limitation is the 2D spatial treatment of the actual 3D world. It is worth noting however, that our models (PIX and REG) and setting trivially generalize to 3D if appropriate data are available. 

\subsection*{Acknowledgments}
This work has been supported by the CHIST-ERA EU project MUSTER\footnote{http://www.chistera.eu/projects/muster} and by the KU Leuven grant RUN/15/005. G.C. additionally acknowledges a grant from the IV\&L network\footnote{http://ivl-net.eu} (ICT COST Action IC1307).

\bibliography{AAAI_biblio}
\bibliographystyle{aaai}

\end{document}